\documentclass[10pt]{article}

\usepackage[a4paper,margin=1in]{geometry}
\usepackage{times}
\usepackage{graphicx}
\usepackage{booktabs}
\usepackage{amsmath,amssymb}
\usepackage{xcolor}
\usepackage{hyperref}
\usepackage[numbers,sort&compress]{natbib}

\usepackage{makecell}

\hypersetup{
    colorlinks=true,
    linkcolor=blue,
    citecolor=blue,
    urlcolor=magenta
}

\title{DriveCtrl: Conditioned Sim-to-Real Driving Video Generation}

\author{
\begin{tabular}{c}
Haonan Zhao$^{1,*}$ \quad
Yiting Wang$^{1,*}$ \quad
Jingkun Chen$^{2}$\\
Valentina Donzella$^{3}$ \quad
Thomas Bashford-Rogers$^{1}$ \quad
Kurt Debattista$^{1}$\\[0.6em]
{\small $^{1}$University of Warwick}\\
{\small $^{2}$Northwestern Polytechnical University}\\
{\small $^{3}$Queen Mary University of London}\\[0.3em]
{\scriptsize $^{*}$Equal contribution}
\end{tabular}
}

\date{}

\begin{document}
\maketitle

\begin{abstract}
Large-scale labelled driving video data is essential for training autonomous driving systems. Although simulation offers scalable and fully annotated data, the domain gap between synthetic and real-world driving videos significantly limits its utility for downstream deployment. Existing video generation methods are not well-suited for this task, as they fail to simultaneously preserve scene structure, object dynamics, temporal consistency, and visual realism, all of which are critical for maintaining annotation validity in generated data. In this paper, we present DriveCtrl, a depth-conditioned controllable sim-to-real video generation framework for realistic driving video synthesis. Built upon a pretrained video foundation model, DriveCtrl introduces a structure-aware adapter that enables depth-guided generation while preserving the scene layout and motion patterns of the source simulation, producing temporally coherent driving videos that remain aligned with the original simulated sequences. We further introduce a scalable data generation pipeline that transforms simulator videos into realistic driving footage matching the visual style of a target real-world dataset. The pipeline supports three conditioning signals: structural depth, reference-dataset style, and text prompts, while preserving frame-level annotations for downstream perception tasks. To better assess this task, we propose a driving-domain-specific knowledge-informed evaluation metric called Driving Video Realism Score (DVRS) that holistically assesses generated videos across four dimensions: plausibility, temporal coherence, structural consistency, and perceptual quality. These dimensions are then integrated into a unified realism score. Experiments demonstrate that DriveCtrl consistently outperforms the base model and competing alternatives in realism, temporal quality, and perception task performance, substantially narrowing the sim-to-real gap for driving video generation.

\end{abstract}

\section{Introduction}

Large-scale labelled driving video data is essential for training and evaluating autonomous driving systems. Recent progress in perception, prediction, and planning has been driven by large real-world benchmarks with rich annotations~\cite{cordts2016cityscapes,yu2020bdd100k,sun2020waymo,caesar2020nuscenes}. However, collecting driving videos at scale remains costly. Rare but safety-critical scenarios are difficult to capture, and dense frame-wise annotation over long temporal sequences is highly labour-intensive \cite{zhao2024exploring}. These limitations make it difficult to build large-scale labelled video datasets for autonomous driving in a scalable and affordable way.

\begin{figure}[!t]
    \centering
    \includegraphics[width=0.47\textwidth]{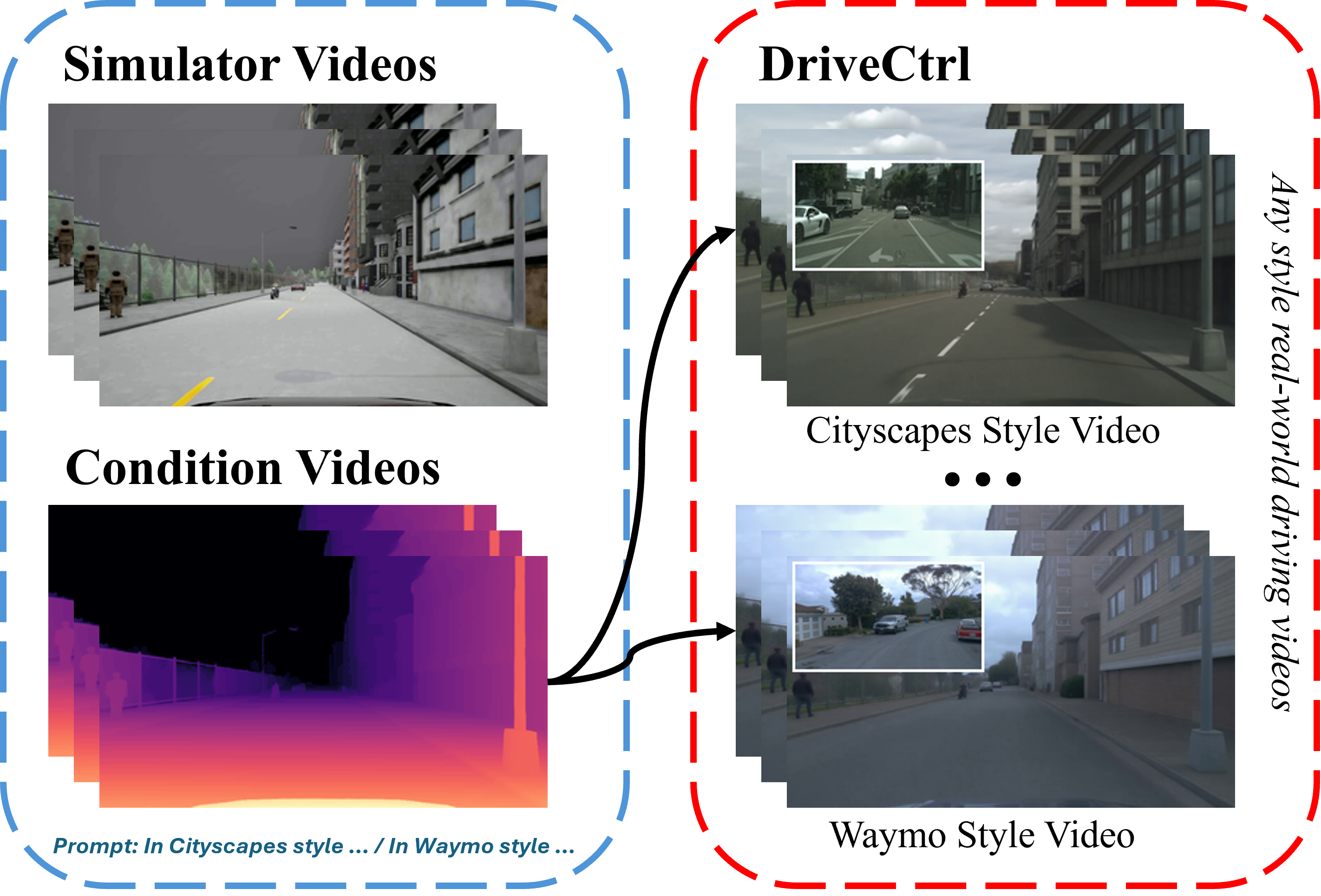}
    \caption{Sim-to-real driving video generation. Given a clip of depth map video, one real-world image and a simple prompt, DriveCtrl generates high-quality driving videos with any corresponding real-world driving dataset style. }
    \label{fig:intro}
\end{figure}

Simulation offers an attractive alternative. Modern driving simulators can generate large amounts of data with accurate labels and fine-grained control over scene layout, traffic participants, weather, and camera settings. Synthetic driving datasets~\cite{sun2022shift,kar2019metasim} have shown the value of simulation for scalable data generation and controlled studies of domain shift. Nevertheless, simulator videos still differ substantially from real-world driving data in appearance, dynamics, and visual realism, which limits their direct usefulness for real-world deployment. Reducing this sim-to-real gap while preserving the original annotations therefore remains a key challenge.

Recent advances in video generation have made it increasingly promising to translate simulated videos into realistic ones. General-purpose video generation and controllable video synthesis models have demonstrated impressive capabilities in visual quality, temporal modelling, and conditional generation~\cite{zhang2024controlvideo,guo2023animatediff,yang2024cogvideox}. However, generic text-to-video and controllable video synthesis models are not well-suited for autonomous driving scenarios, where generated content must remain faithful to source scene structure and object dynamics to preserve annotation validity across frames. Moreover, none of these methods explicitly targets the practical goal of scalable sim-to-real video translation with reusable frame-wise annotations.

Driving-specific generative models~\cite{gao2024magicdrive,li2024drivingdiffusion,lu2024wovogen,wang2024drivedreamer,zhao2025drivedreamer2,wang2024drivewm,mao2025dreamdrive} have also made notable progress by leveraging richer control signals such as camera trajectories, road layouts, 3D boxes, map priors, multi-view observations, and world representations. While these methods improve realism and controllability, they are mainly designed for scene synthesis, world modelling, or future generation, rather than scalable sim-to-real video translation with reusable frame-wise labels. These limitations highlight the need for a lightweight, annotation-preserving sim-to-real translation approach that prioritises structural fidelity, temporal coherence, and visual realism without requiring heavy conditioning infrastructure.

To this end, we propose \textbf{\textit{DriveCtrl}}, a depth-conditioned sim-to-real driving video generation framework. Built upon a pretrained video foundation model, DriveCtrl introduces a structure-aware adapter that enables depth-guided video generation while preserving the scene layout and motion patterns of the source simulation. This allows the generated sequences to become more realistic and temporally coherent without breaking structural alignment with the original simulator videos. Building on this, we develop a scalable annotation-preserving data generation pipeline that converts simulator videos into realistic driving footage matching the visual style of any target real-world dataset. The pipeline takes three complementary conditioning inputs, namely depth maps for structural control, reference images for style guidance, and text prompts for semantic control, enabling flexible and controllable synthesis while preserving the original frame-wise annotations.

Evaluation is another important challenge. Feature-distribution metrics such as FID~\cite{heusel2017gans} and FVD~\cite{unterthiner2018fvd} measure visual fidelity but are agnostic to driving-domain semantics. They cannot assess whether a generated scene is physically plausible, temporally coherent, or consistent with real-world driving knowledge. To address this gap, we further propose a driving-domain-specific world-knowledge-informed metric, termed \textbf{\textit{Driving Video Realism Score (DVRS)}}, which evaluates generated videos from the aspects of plausibility, coherence, consistency, and visual realism to form an overall realism score. Combined with standard video-generation and downstream perception evaluations, DVRS provides a more task-relevant assessment of whether sim-to-real driving video generation truly narrows the gap to real-world data.

Our contributions are summarised as follows:
\begin{itemize}
    \item We propose \textbf{\textit{DriveCtrl}}, a depth-conditioned sim-to-real driving video generation framework that introduces a structure-aware adapter for pretrained video foundation models, improving realism and temporal coherence while preserving scene structure and motion patterns.
    
    \item We develop a scalable annotation-preserving data generation pipeline that transforms simulator videos into realistic driving videos in arbitrary real-world dataset styles under structural control, reference-dataset style guidance, and prompt conditioning, enabling scalable construction of realistic labelled driving video datasets.
    
    \item We propose \textbf{\textit{Driving Video Realism Score (DVRS)}}, a driving-domain-specific world-knowledge-informed evaluation metric, and show through extensive experiments that DriveCtrl consistently improves realism, video quality, and downstream perception performance, substantially narrowing the sim-to-real gap for driving video generation.
\end{itemize}

\begin{figure}[!t]
    \centering
    \includegraphics[width=1\textwidth]{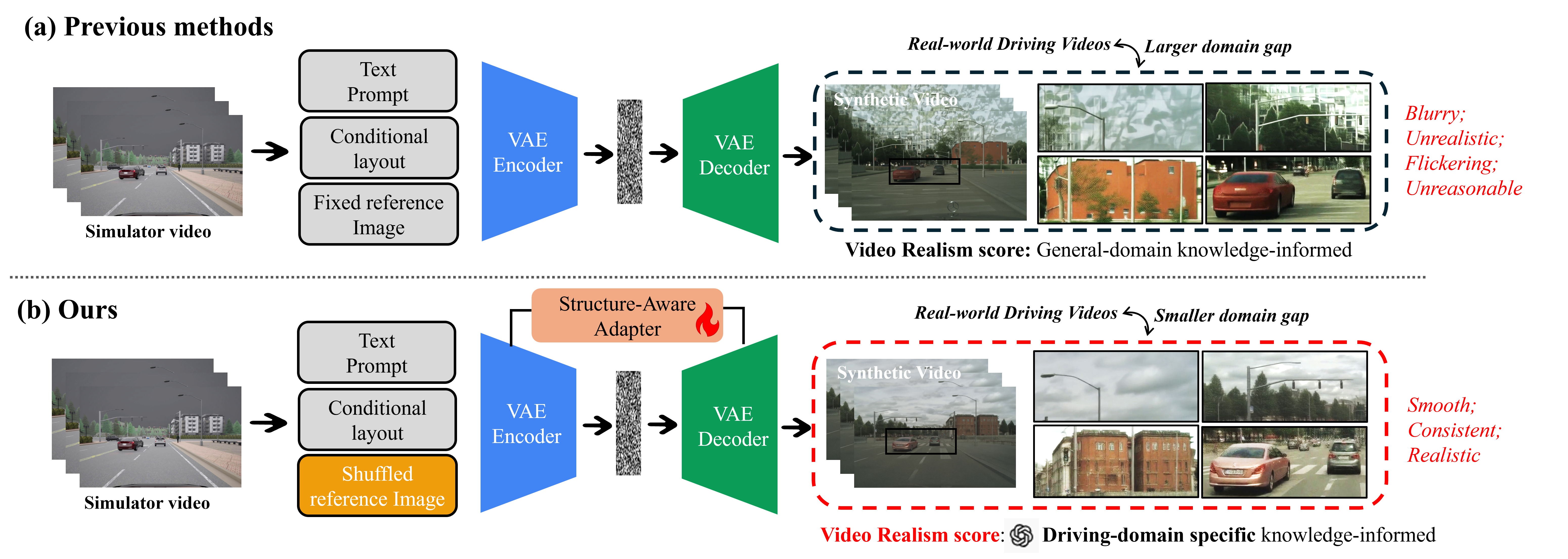}
    \caption{Comparison between previous video synthesis approaches and the proposed DriveCtrl.}
    \label{fig:pipeline_small}
\end{figure}

\begin{figure*}[!htbp]
    \centering
    \includegraphics[width=1\textwidth]{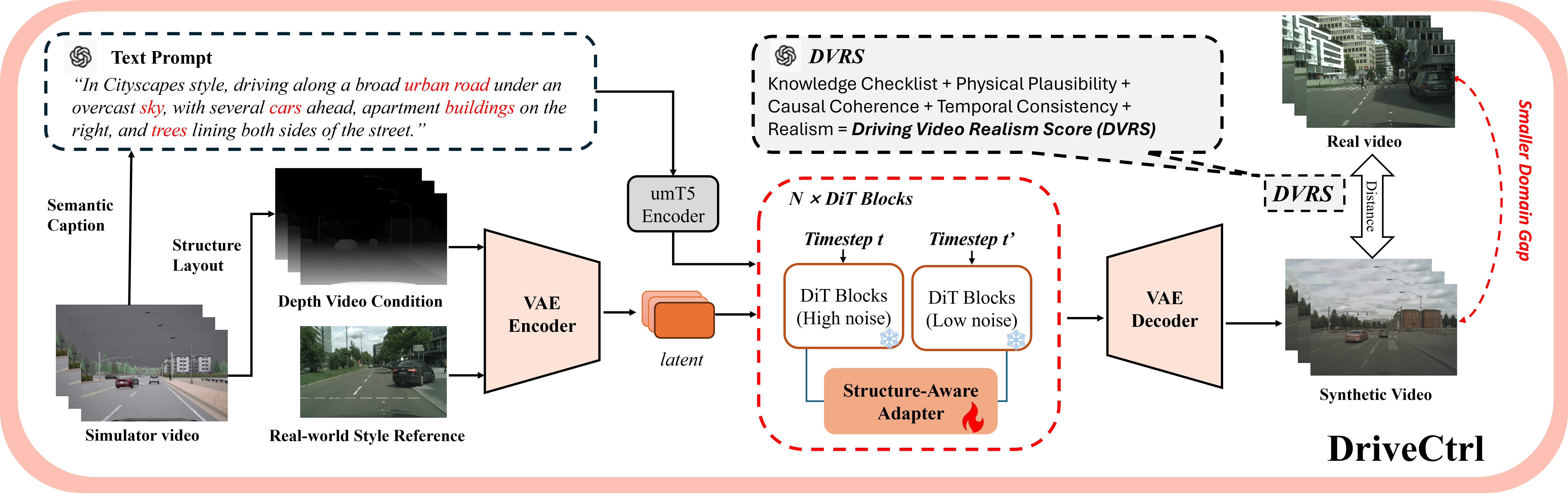}
    \caption{Overview of the proposed DriveCtrl framework. Given a condition video, a text prompt, and a reference image sampled from the target real-world domain, DriveCtrl injects a structure-aware adapter into a pretrained video foundation model to enable realistic driving video generation. Realism is further assessed by the proposed Driving Video Realism Score (DVRS)}
    \label{fig:pipeline_main}
\end{figure*}

\section{Related Work}

\subsection{Video Generation}
Generating temporally coherent and visually realistic videos from structured conditions is central to our sim-to-real translation goal. Diffusion-based video foundation models~\cite{ho2022videodiffusion,yang2024cogvideox,wan2025wan} have demonstrated strong open-domain synthesis capabilities in both text-to-video and image-to-video settings. Building on these, controllable generation methods introduce explicit conditioning signals to guide the synthesis process. Control-A-Video and ControlVideo leverage structural conditions such as depth and edge maps~\cite{chen2023controlavideo,zhang2024controlvideo}, whereas DragNUWA and MotionCtrl incorporate trajectory and camera motion priors to achieve finer-grained spatial control~\cite{yin2023dragnuwa,wang2024motionctrl}. Style-conditioned and reference-image-guided generation methods~\cite{ye2023ip} have also explored appearance transfer from reference images, yet they focus on subject or portrait consistency rather than driving-scene structural fidelity. However, these methods optimise for perceptual quality in open-domain settings and lack mechanisms to enforce structural alignment between input conditions and generated frames — a critical requirement when generated videos must inherit frame-wise annotations from simulation. More recently, Cosmos-Transfer~\cite{nvidia2025cosmostransfer} and OmniVDiff~\cite{xi2025omnivdiff} have demonstrated strong multi-condition video generation quality, but they are designed for high-fidelity general synthesis rather than structure-preserving sim-to-real translation, and their substantial computational requirements further limit scalability for large-scale labelled data generation.

\subsection{Driving Data Synthesis and Sim-to-Real Translation}
Synthetic driving datasets including Synscapes, Meta-Sim, and SHIFT~\cite{wrenninge2018synscapes,kar2019metasim,sun2022shift} demonstrate the value of simulation for scalable data generation and controlled evaluation. Compared with real-world data collection, simulation-based approaches offer unlimited data volume, automatic and accurate annotation, and precise control over environmental factors. Complementary efforts explore neural scene reconstruction~\cite{yang2023unisim} and geometric-controlled data generation~\cite{chen2024geodiffusion}, though neither targets annotation-preserving video-level sim-to-real translation. To reduce the sim-to-real gap, prior works have explored realistic sensor synthesis and appearance translation, including SurfelGAN, Enhancing Photorealism Enhancement, and CARLA2Real~\cite{yang2020surfelgan,richter2021epe,pasios2025carla2real}. Video-to-video translation methods~\cite{mallya2020world,liang2024flowvid} have also explored domain adaptation at the video level, but they rely on paired training data and struggle to maintain temporal consistency at scale. These methods improve realism in isolation but are not designed to generate photorealistic driving videos that preserve simulation structure for direct frame-wise annotation reuse.

Recent driving-specific generative models have made significant progress in controllable scene synthesis. Methods such as MagicDrive, DrivingDiffusion, and WoVoGen~\cite{gao2024magicdrive,li2024drivingdiffusion,lu2024wovogen} leverage structured inputs including HD maps, 3D bounding boxes, and camera poses to generate realistic street-view images or multi-view videos. Panacea~\cite{wen2024panacea} and Panacea+~\cite{wen2024panaceaplus} further address panoramic multi-view video generation through decomposed 4D attention and BEV-layout-based conditioning. MagicDrive3D~\cite{gao2024magicdrive3d} extends this line of work to 3D scene generation by combining multi-view video synthesis with 3D Gaussian Splatting. In parallel, driving world models~\cite{hu2023gaia1,wang2024drivedreamer,gao2024vista,wang2024drivewm,zhao2025drivedreamer2,mao2025dreamdrive} model the temporal evolution of driving scenes conditioned on trajectories, actions, or multi-view observations for future prediction and planning. While these works represent important progress in scene synthesis and world modelling, they rely on heavyweight control signals such as HD maps, 3D bounding boxes, and multi-view camera poses that are expensive to obtain outside of dedicated simulation pipelines, limiting their applicability as general-purpose sim-to-real data generation tools.

\subsection{Realism Evaluation Metrics}
While FID~\cite{heusel2017gans} and FVD~\cite{unterthiner2018fvd} measure distributional similarity at the feature level, they are agnostic to semantic plausibility and scene-level realism. Beyond these single-score metrics, comprehensive benchmarks such as VBench~\cite{huang2024vbench,zheng2025vbench2} evaluate video generation quality from multiple perspectives including fidelity, motion, and consistency. More recently, world-knowledge-informed evaluation methods~\cite{bansal2024videophy,meng2024towards,niu2025wise,zhang2025worldgenbench,bansal2025videophy,chen2026t2vworldbench} have emphasised that generative outputs should also be assessed for physical commonsense, semantic plausibility, and alignment with real-world knowledge by leveraging LLM- or VLM-based evaluators. However, existing world-knowledge-informed metrics are designed for general video generation and do not account for driving-domain-specific factors such as traffic dynamics, road geometry, and scene plausibility — motivating our proposed \textbf{\textit{DVRS}}, which is tailored to the structural and semantic requirements of realistic driving video evaluation.

\section{Methodology}

Our method consists of two components. First, we propose a \emph{structure-aware adapter} that adapts a pretrained video generation model for condition-video-guided sim-to-real driving video generation. Second, based on the adapted generator, we build a \emph{sim-to-real driving video data generation pipeline} that converts simulator video datasets into photorealistic driving datasets while preserving the original frame-wise annotations.

\subsection{Structure-Aware Adapter}

\textbf{Overview.} As shown in Fig. \ref{fig:pipeline_main}, we build a structure-aware adapter on top of a pretrained video foundation model~\cite{wan2025wan} to adapt it for sim-to-real driving video generation. Given a structural condition video $C$, a text prompt $p$, and a style reference image $I^r$, the goal is to generate a photorealistic driving video $V$ that preserves the scene layout and motion encoded in $C$ while using $I^r$ only as weak appearance guidance. In the original model, the reference image is typically the first frame of the target video and is used to guide the generation of subsequent frames. In contrast, in our setting, the reference image is intentionally decoupled from the target video, serving only as an unpaired style reference. The key idea is to explicitly bias the model toward condition-video-guided generation: the condition video should dominate geometry and temporal dynamics, whereas the reference image should only contribute visual style.  

\noindent\textbf{Adapter design.} Our adapter is attached to the pretrained foundation model backbone~\cite{wan2025wan}, which adopts a flow-matching formulation for video generation, while keeping the original backbone frozen. The structural condition video is injected as the primary condition before patch embedding, so that its geometric and temporal information is preserved throughout denoising. The reference image is still provided through the original image-conditioning pathways of the backbone. To efficiently adapt the model, we implement the proposed adapter with lightweight low-rank updates~\cite{hu2022lora} inserted into the transformer blocks. Specifically, for each DiT block, LoRA is injected into the four projection layers of self-attention, the four projection layers of cross-attention, and the two linear layers of the feed-forward network. The adapted model can be written as
\begin{equation}
\hat{u}_t = \mathcal{F}_{\Theta,\Delta_b}(x_t, C, p, I^{r}, t),
\end{equation}
where $\Theta$ denotes the frozen pretrained backbone, $\Delta_b$ denotes the trainable adapter parameters for branch $b \in \{\mathrm{high}, \mathrm{low}\}$, $x_t$ is the noisy latent at timestep $t$, and $\hat{u}_t$ is the predicted flow-matching target.

\noindent\textbf{Random reference-image conditioning.} The reference image may leak spatially aligned structural priors to the generator, causing the model to rely on the reference image rather than the condition video for scene layout. This is because the original backbone is trained to regard the reference image as the first frame and to generate the remaining frames conditioned on it. To address this, we deliberately randomise the reference-image assignment during training. For the $i$-th video segment in the training set, its original reference image $I_i^r$, which is also the first frame of this segment, is replaced with a randomly assigned one $\tilde{I}_i^r$. This operation breaks the spatial correspondence between the reference image and the target video, preventing the reference image from being interpreted as the first frame. Instead, the model is encouraged to rely on the structural condition video for scene layout and motion, while using the reference image mainly for visual appearance cues. The resulting generation process can therefore be expressed as

\begin{equation}
\hat{V}_i = \mathcal{G}_{\Theta,\Delta}(C_i, p_i, \tilde{I}_i^r),
\end{equation}
where $\mathcal{G}_{\Theta,\Delta}$ denotes the full denoising 
process parameterized by frozen backbone $\Theta$ and trainable adapter $\Delta$, and $\hat{V}_i$ denotes the generated video. The same random reference-image conditioning is used in both training and inference.

\noindent\textbf{Training objective.} Following the original design, we use a high-noise checkpoint and a low-noise checkpoint for different denoising stages. At inference time, they are switched according to the current timestep. We retain the original flow matching objective of the pretrained backbone and optimise only the adapter parameters. Let $u_t^\star$ denote the target velocity at timestep $t$. For each checkpoint $b \in \{\mathrm{high}, \mathrm{low}\}$, the loss is
\begin{equation}
\mathcal{L}_{b}
=
\mathbb{E}_{(V,C,p,\tilde{I}^{r}),\, t\sim \mathcal{U}(\mathcal{T}_b)}
\left[
\left\|
\mathcal{F}_{\Theta,\Delta_b}(x_t, C, p, \tilde{I}^{r}, t) - u_t^\star
\right\|_2^2
\right],
\end{equation}
where $\mathcal{T}_b$ denotes the timestep interval assigned to checkpoint $b$. The total loss is the sum of the two stage-specific objectives:
\begin{equation}
\mathcal{L} = \mathcal{L}_{\mathrm{high}} + \mathcal{L}_{\mathrm{low}}.
\end{equation}

\subsection{Sim-to-Real Driving Video Generation Pipeline}

Building upon the proposed structure-aware adapter, we further introduce a sim-to-real driving video generation pipeline for cost-effective construction of photorealistic annotated driving video datasets. For each simulator video clip, we first extract a structural condition video, pair it with a text prompt and a randomly sampled, unpaired reference image from a real driving dataset, and then feed them into the adapted video generator. Although the framework is compatible with different modalities of inputs, we use depth-map videos in this work because depth provides a geometry-aware control signal that preserves scene layout and motion structure while being less affected by visual appearance. This makes depth particularly suitable for sim-to-real video generation. Specifically, we use depth-map videos predicted from the simulator sequence as condition videos, rather than directly using existing depth from the simulator. This is to demonstrate that the proposed pipeline can still achieve effective sim-to-real video generation even when simulator-provided depth is unavailable. In addition, this choice does not compromise the validity of the experimental evaluation.

\begin{figure}[!h]
    \centering
    \includegraphics[width=0.8\textwidth]{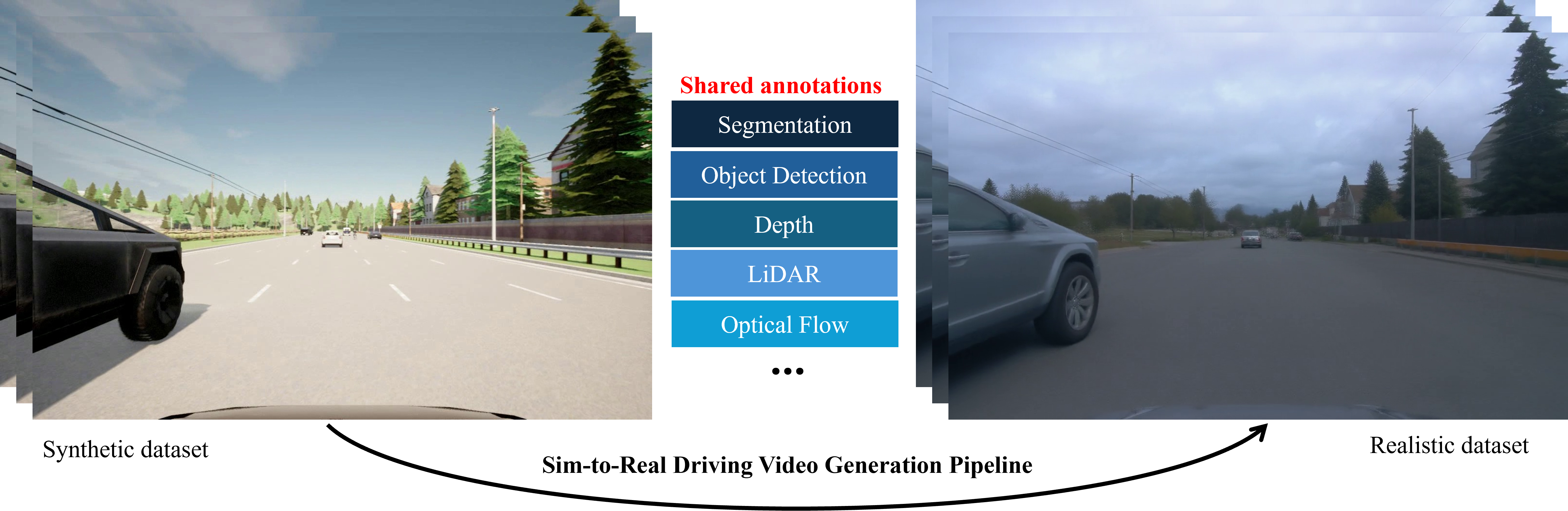}
    \caption{Illustration of annotation sharing between simulated and realistic driving datasets, where the translated videos preserve frame-wise annotations.}
    \label{fig:sim2real}
\end{figure}

Because the generator is optimised for condition-guided generation, the translated video largely preserves the scene layout and motion patterns of the source simulator clip, while the text prompt and the random reference image provide complementary semantic and appearance guidance. As a result, the generated video becomes visually closer to real driving data while remaining approximately aligned with the original simulator sequence. As Fig.~\ref{fig:sim2real} shows, this property makes it practical to reuse the original frame-wise annotations for the translated videos, enabling low-cost construction of a more realistic annotated driving video dataset.

\subsection{Driving Video Realism Score (DVRS)}

To quantify the realism of generated driving videos, we introduce \emph{Driving Video Realism Score (DVRS)}, an LLM-based evaluation metric specifically designed for autonomous-driving video data. Inspired by T2VWorldBench~\cite{chen2026t2vworldbench} and PhyGenBench~\cite{meng2024towards}, which leverage LLM- or VLM-based evaluators to assess world knowledge and physical commonsense beyond low-level feature similarity, DVRS evaluates how far generated driving videos deviate from real driving data under perspectives such as traffic knowledge, common sense and physical plausibility.

\noindent\textbf{Driving-specific checklist construction.}
Rather than using a generic prompt-alignment evaluation, we first construct a driving-specific checklist tailored to ego-view urban driving. Following the structured evaluation philosophy of WorldGenBench~\cite{zhang2025worldgenbench}, the evaluator is prompted with a scene prototype describing front-facing urban driving videos with roads, vehicles, pedestrians, buildings, traffic lights, road signs, and lane markings, and is asked to generate a set of concrete, visually checkable realism criteria. These items are constrained to be short, objective, and grounded in observable traffic behaviour, physics, and social conventions. Let $\mathcal{Q}=\{q_1,\dots,q_M\}$ denote the resulting checklist. In practice, the checklist is generated once and reused for all evaluated datasets to ensure consistency across comparisons.

\noindent\textbf{Checklist and realism scoring.}
Given a driving segment $s$ and the checklist $\mathcal{Q}$, the multimodal judge returns five scores corresponding to the knowledge checklist score $R_{\mathrm{kc}}(s)$, physical plausibility $R_{\mathrm{phy}}(s)$, causal coherence $R_{\mathrm{cau}}(s)$, temporal consistency $R_{\mathrm{tmp}}(s)$, and visual realism $R_{\mathrm{vis}}(s)$:
\begin{equation}
\big(R_{\mathrm{kc}}(s),\, R_{\mathrm{phy}}(s),\, R_{\mathrm{cau}}(s),\, R_{\mathrm{tmp}}(s),\, R_{\mathrm{vis}}(s)\big)
=
\mathcal{E}_{\mathrm{LLM/VLM}}(s,\mathcal{Q}),
\end{equation}
where $\mathcal{E}_{\mathrm{LLM/VLM}}(s,\mathcal{Q})$ denotes the evaluation function implemented by an LLM- or VLM-based judge. In particular, $R_{\mathrm{kc}}(s)$ is computed as the proportion of checklist items marked as satisfied. This follows the same high-level spirit as the knowledge checklist score in WorldGenBench~\cite{zhang2025worldgenbench}, but is redesigned for autonomous-driving videos rather than generic text-to-image outputs. The checklist score measures how many expected realism properties of urban driving are explicitly satisfied by the segment. 

To capture realism beyond checklist satisfaction, we additionally evaluate dynamic realism and visual realism at the dataset level. These evaluations are performed by prompting an LLM/VLM-based judge to assess each segment under multiple realism dimensions. Specifically, dynamic realism is characterised by physical plausibility, causal coherence, and temporal consistency, while visual realism is measured separately as an overall perceptual realism score. Physical plausibility evaluates whether motion and geometry obey basic physical principles, causal coherence evaluates whether traffic events follow reasonable cause-and-effect patterns, temporal consistency evaluates whether scene evolution remains coherent over time, and visual realism captures the judge's overall impression of whether the segment visually resembles a real driving video. This design is conceptually related to recent multi-dimensional evaluation protocols~\cite{niu2025wise,chen2026t2vworldbench} for generative models, while remaining adapted to driving video realism. Let $\bar{R}_{x}(\mathcal{S})$ denote the average score of component $x$ over all segments in dataset $\mathcal{S}$. We then compute the absolute distance between the generated and real datasets for each component:
\begin{equation}
D_{x}
=
\left|
\bar{R}_{x}(\mathcal{S}_{\mathrm{gen}})
-
\bar{R}_{x}(\mathcal{S}_{\mathrm{real}})
\right|,
\quad
x\in\{\mathrm{kc},\mathrm{phy},\mathrm{cau},\mathrm{tmp},\mathrm{vis}\}.
\end{equation}

\noindent\textbf{Driving Video Realism Score.}
We first define the dynamic realism distance as the mean of the physical plausibility, causal coherence, and temporal consistency distances:
\begin{equation}
D_{\mathrm{dyn}}
=
\operatorname{Avg}\!\left(
D_{\mathrm{phy}},
D_{\mathrm{cau}},
D_{\mathrm{tmp}}
\right).
\end{equation}
The final DVRS is then defined as:
\begin{equation}
\mathrm{DVRS}
=
\lambda_{\mathrm{kc}}\,D_{\mathrm{kc}}
+\lambda_{\mathrm{dyn}}\,D_{\mathrm{dyn}}
+\lambda_{\mathrm{vis}}\,D_{\mathrm{vis}}.
\label{eq:dvrs}
\end{equation}
Here, $\lambda$ denotes the weight assigned to each corresponding component. A smaller DVRS indicates that the generated videos are closer to real driving data under a world-knowledge-informed realism criterion.

\begin{figure*}[!t]
    \centering
    \includegraphics[width=0.7\textwidth]{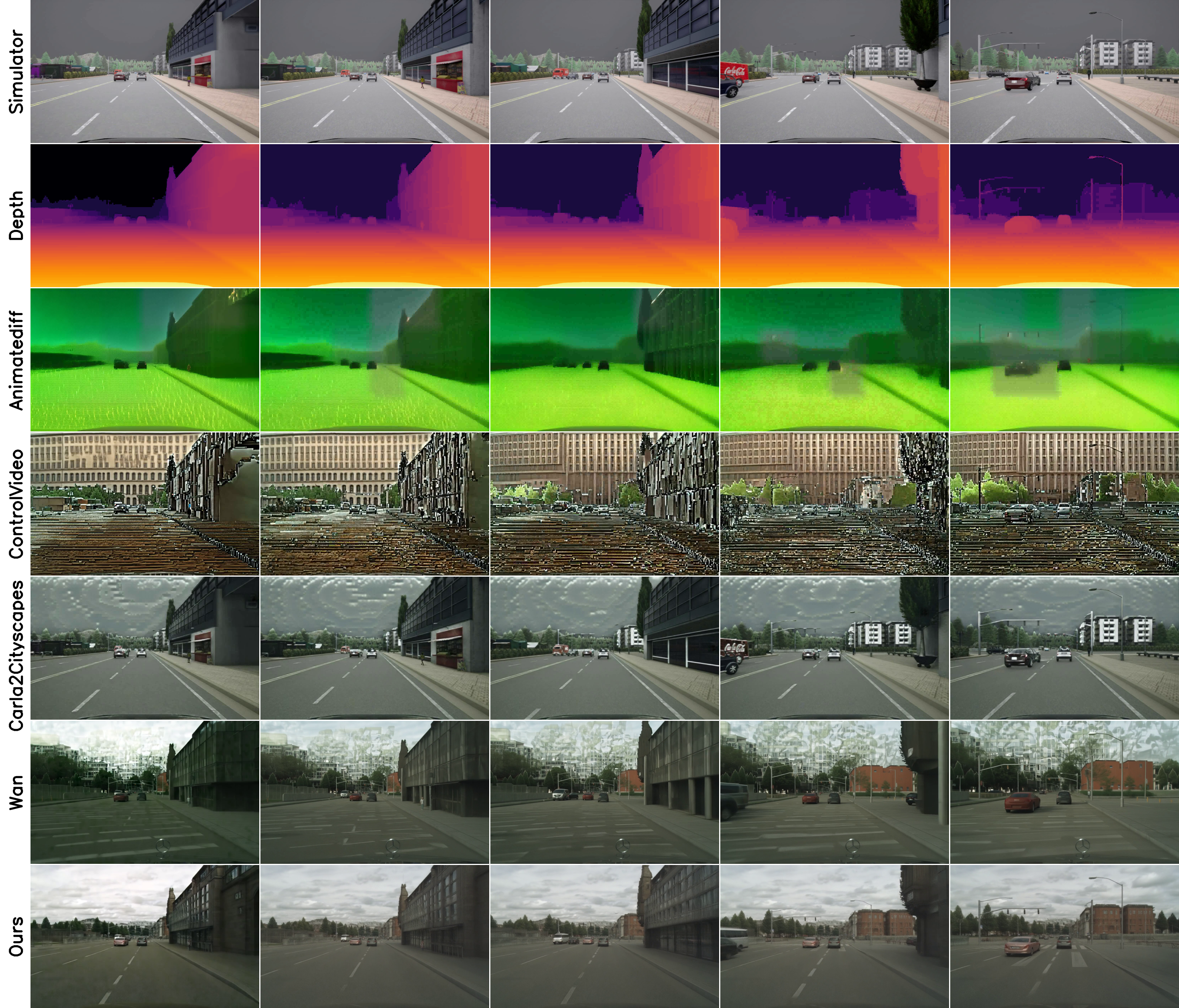}
    \caption{Qualitative comparison on the Cityscapes dataset}
    \label{fig:comparison_cityscapes}
\end{figure*}

\begin{figure*}[!t]
    \centering
    \includegraphics[width=0.7\textwidth]{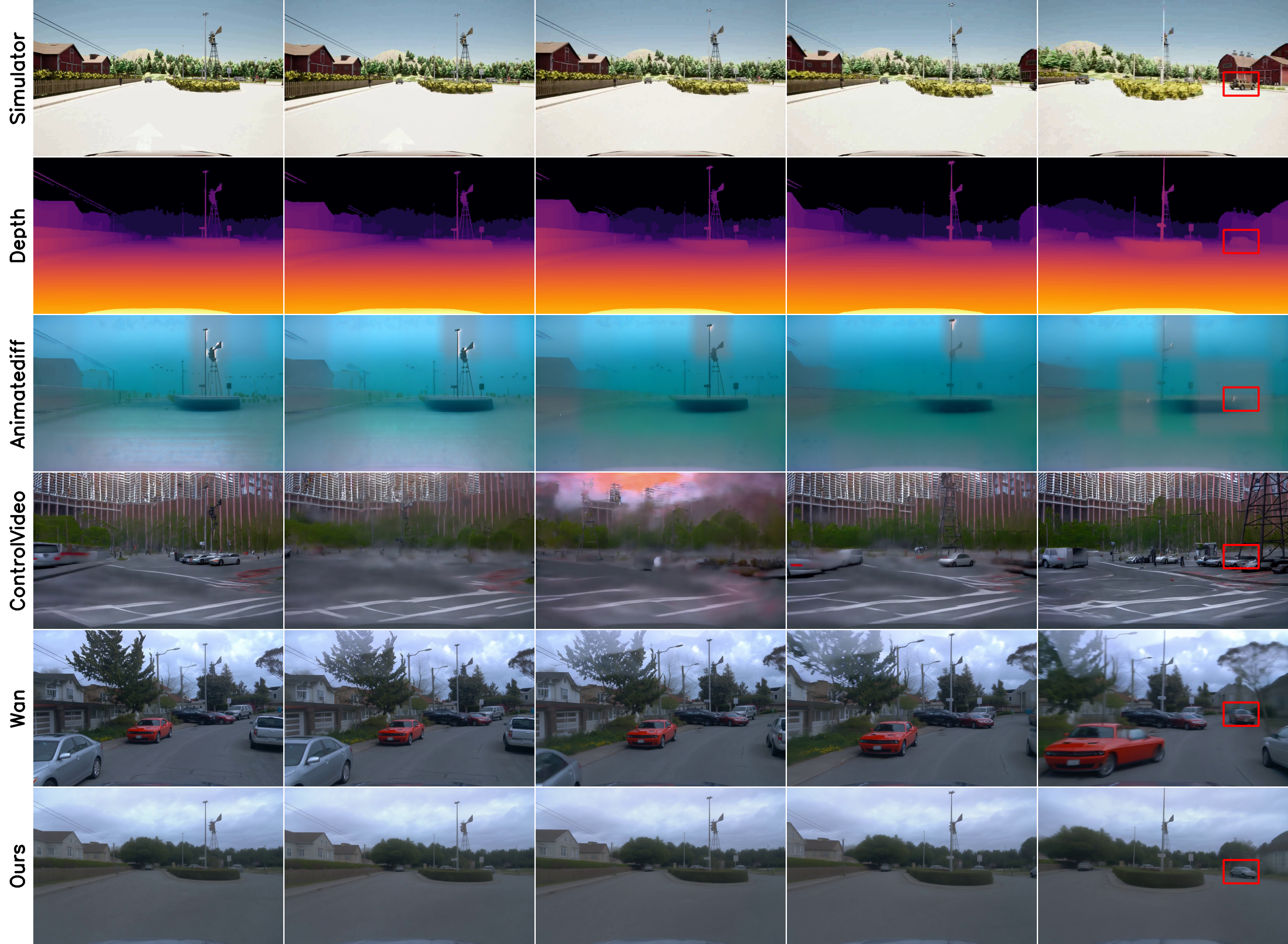}
    \caption{Qualitative comparison on the Waymo dataset}
    \label{fig:comparison_waymo}
\end{figure*}

\begin{table}[!h]
\centering
% \small
% \scriptsize
% \setlength{\tabcolsep}{3.5pt}
% \renewcommand{\arraystretch}{0.85}
\resizebox{\textwidth}{!}{%

\begin{tabular}{lcccccc}
\toprule
Method 
& Knowledge Checklist $\downarrow$
& Physical Plausibility $\downarrow$
& Causal Coherence $\downarrow$
& Temporal Consistency $\downarrow$
& Visual Realism $\downarrow$
& DVRS $\downarrow$ \\
\midrule
\textbf{Cityscapes} intra-domain
& 1.11
& 0.03
& 0.07
& 0.24
& 0.74
& 0.99 \\
Animatediff~\cite{guo2023animatediff}
& 41.22
& 4.80
& 3.30
& 3.20
& 40.20
& 39.70 \\
ControlVideo~\cite{zhang2024controlvideo}
& 62.22
& 6.58
& 5.00
& 5.80
& 61.05
& 60.40 \\
Carla2Cityscapes~\cite{pasios2025carla2real}
& 5.55
& 2.45
& 1.37
& 2.23
& 20.92
& 15.55 \\
Wan~\cite{wan2025wan}
& 5.33
& 1.80
& 1.28
& 2.20
& 17.33
& 13.42 \\
Ours
& \textbf{3.55}
& \textbf{1.47}
& \textbf{0.78}
& \textbf{2.07}
& \textbf{13.93}
& \textbf{10.63} \\
\cmidrule(lr){1-7}
\textbf{Waymo} intra-domain
& 2.89
& 0.30
& 0.33
& 0.47
& 3.47
& 3.34 \\
Animatediff~\cite{guo2023animatediff}
& 26.00
& 4.68
& 2.82
& 3.05
& 37.66
& 32.94 \\
ControlVideo~\cite{zhang2024controlvideo}
& 33.55
& 5.98
& 3.92
& 5.66
& 54.53
& 46.65 \\
Wan~\cite{wan2025wan}
& 11.45
& 2.82
& 1.37
& 2.17
& 22.09
& 18.25 \\
Ours
& \textbf{4.89}
& \textbf{1.70}
& \textbf{0.98}
& \textbf{1.18}
& \textbf{13.64}
& \textbf{10.47} \\
\bottomrule
\end{tabular}%
}
\caption{DVRS evaluation results on the Cityscapes and Waymo datasets.} 
\label{tab:dvrs_combined}
\end{table}

\begin{table}[!h]
\centering
% \small
% \setlength{\tabcolsep}{6pt}
% \renewcommand{\arraystretch}{0.85}
\resizebox{\textwidth}{!}{%
\begin{tabular}{lcccc}
\toprule
Method 
& Background Consistency $\uparrow$
& Subject Consistency $\uparrow$
& Motion Smoothness $\uparrow$
& Temporal Flickering $\uparrow$ \\
\midrule
\multicolumn{5}{c}{\textbf{Cityscapes}} \\
\cmidrule(lr){1-5}
Animatediff~\cite{guo2023animatediff}
& 0.954
& 0.941
& 0.963
& 0.956 \\
ControlVideo~\cite{zhang2024controlvideo}
& 0.899
& 0.878
& 0.831
& 0.817 \\
Carla2Cityscapes~\cite{pasios2025carla2real}
& 0.928
& 0.933
& 0.986
& 0.973 \\
Wan~\cite{wan2025wan}
& \textbf{0.964}
& 0.964
& 0.992
& 0.980 \\
Ours
& \textbf{0.964}
& \textbf{0.966}
& \textbf{0.993}
& \textbf{0.987} \\
\cmidrule(lr){1-5}
\multicolumn{5}{c}{\textbf{Waymo}} \\
\cmidrule(lr){1-5}
Animatediff~\cite{guo2023animatediff}
& 0.942
& 0.929
& 0.968
& 0.961 \\
ControlVideo~\cite{zhang2024controlvideo}
& 0.924
& 0.868
& 0.967
& 0.959 \\
Wan~\cite{wan2025wan}
& \textbf{0.950}
& 0.957
& 0.991
& 0.980 \\
Ours
& \textbf{0.950}
& \textbf{0.961}
& \textbf{0.994}
& \textbf{0.990} \\
\bottomrule
\end{tabular}%
}
\caption{VBench evaluation results on the Cityscapes and Waymo datasets.}
\label{tab:vbench_combined}
\end{table}

\section{Experiments}
\subsection{Experiment Settings}

\noindent\textbf{Implementation Details.}
We fine-tune our model based on Wan~\cite{wan2025wan}, a large-scale video foundation model, adopting Wan2.2-Fun-A14B-Control as the backbone. We train two lightweight structural control adapters operating at different noise levels: a high-noise adapter that captures coarse scene layout and global structure, and a low-noise adapter that refines fine-grained details and local consistency. Each adapter is trained independently on a single NVIDIA L40 GPU for 25 epochs (12 hours per run).

\noindent\textbf{Datasets.}
Our training set is constructed from two widely used autonomous driving benchmarks: Cityscapes~\cite{cordts2016cityscapes} and Waymo Open Dataset~\cite{sun2020waymo}. For each dataset, we select 10 non-overlapping daytime video clips of 40 frames, paired with depth videos predicted by Depth Anything V2~\cite{yang2024depth}, random reference images, and text prompts as conditioning inputs. For evaluation, we follow our proposed sim-to-real pipeline and use depth videos from the SHIFT dataset~\cite{sun2022shift} as structural input to generate videos in Cityscapes and Waymo styles. Each evaluation set contains 60 videos of 40 frames (2,400 frames total), following evaluation scales comparable to prior works~\cite{guo2023animatediff,zhang2024controlvideo,pasios2025carla2real}. We additionally sample 2,400 frames from Cityscapes and Waymo respectively as real-world reference sets.

\noindent\textbf{Metrics.}
We conduct three types of evaluation. \textbf{(1) Realism:} We use our proposed \textit{DVRS} to assess generated videos across plausibility, temporal coherence, structural consistency, and perceptual quality via world-knowledge-informed evaluation. In computing DVRS (Eq.~\ref{eq:dvrs}), all component weights are set equally ($\lambda_{\mathrm{kc}}=\lambda_{\mathrm{dyn}}=\lambda_{\mathrm{vis}}$); since $D_{\mathrm{dyn}}$ sub-components are scored on $[0,10]$ while others are on $[0,100]$, $D_{\mathrm{dyn}}$ is rescaled by a factor of 10 for comparability. We do not adopt FID~\cite{heusel2017gans} or FVD~\cite{unterthiner2018fvd}, as these feature-distribution metrics are sensitive to scene-content diversity across clips rather than the realism gap we target — even disjoint real-video subsets from the same dataset can yield large FID/FVD distances. \textbf{(2) Video quality:} We use VBench~\cite{huang2024vbench} for fine-grained perceptual quality assessment across multiple dimensions. \textbf{(3) Downstream perception:} For depth preservation, we apply Depth Anything V2~\cite{yang2024depth} to generated videos and compare predictions against the input depth maps to measure geometric fidelity. For semantic segmentation, we train DeepLabV3~\cite{chen2017deeplabv3} on Cityscapes-style generated videos paired with SHIFT semantic labels, and evaluate on the Cityscapes validation set to measure how effectively generated data supports real-world perception.

\noindent\textbf{Baselines.}
We compare DriveCtrl against multiple state-of-the-art methods. AnimateDiff~\cite{guo2023animatediff} and the ControlNet~\cite{zhang2023adding} within ControlVideo~\cite{zhang2024controlvideo} are retrained on Cityscapes and Waymo using corresponding depth-image pairs and prompts. For Carla2Real~\cite{pasios2025carla2real}, we use the official pretrained Carla2Cityscapes weights. All methods follow the same sim-to-real pipeline to generate evaluation datasets with frame-wise annotations.

\subsection{Qualitative Analysis}
% FID score is 131.59 
% FVD score is 94.22

Figures~\ref{fig:comparison_cityscapes} and~\ref{fig:comparison_waymo} provide qualitative comparisons on the Cityscapes-style and Waymo-style generation settings, respectively. As shown in Fig.~\ref{fig:comparison_cityscapes}, Animatediff~\cite{guo2023animatediff} generally preserves the overall scene structure, but the generated videos suffer from poor color rendering and unrealistic material appearance, making them still look far from real Cityscapes videos. In contrast, ControlVideo~\cite{zhang2024controlvideo} fails to maintain the correct scene structure, leading to clear structural errors in the generated results. Carla2Cityscapes~\cite{pasios2025carla2real} and Wan~\cite{wan2025wan} produce videos with better visual appearance than Animatediff~\cite{guo2023animatediff} and ControlVideo~\cite{zhang2024controlvideo}, but both exhibit block-like artifacts, particularly in the sky regions, which degrade overall realism. Compared with these methods, our approach generates videos with well-preserved structure and a visual style that is much closer to real Cityscapes data~\cite{cordts2016cityscapes}. A similar trend can be observed in Fig.~\ref{fig:comparison_waymo}. Animatediff~\cite{guo2023animatediff} still tends to preserve coarse structure but lacks realistic appearance, while ControlVideo~\cite{zhang2024controlvideo} again shows clear structural inconsistency. Although Wan~\cite{wan2025wan} produces Waymo-style results in this example, its generation is noticeably disturbed by the reference image, which introduces incorrect structural content into the output video. This failure case highlights that the original Wan backbone~\cite{wan2025wan} relies on the reference image and therefore breaks the intended scene layout. In contrast, our method alleviates this issue through the proposed structure-aware adapter, allowing the model to better follow the condition video and generate structurally correct results while maintaining a realistic Waymo-like visual style~\cite{sun2020waymo}.

\begin{table}[!t]
\centering
\small
\setlength{\tabcolsep}{3pt}
\renewcommand{\arraystretch}{0.85}
\begin{tabular}{lcccccc}
\toprule
Method & SSIM $\uparrow$ & AbsRel $\downarrow$ & RMSE $\downarrow$ & $\delta_1$ $\uparrow$ & $\delta_2$ $\uparrow$ & $\delta_3$ $\uparrow$ \\
\midrule
ControlVideo~\cite{zhang2024controlvideo}      & 0.732 & 0.456 & 19.414 & 0.574 & 0.725 & 0.802 \\
Animatediff~\cite{guo2023animatediff}       & \textbf{0.856} & 0.222 & 11.358 & 0.755 & 0.859 & 0.906 \\
Carla2Cityscapes~\cite{pasios2025carla2real}  & 0.852 & 0.290 & 9.862 & 0.771 & 0.849 & 0.891 \\
Wan~\cite{wan2025wan}               & 0.777 & 0.268 & 12.047 & 0.737 & 0.867 & 0.914 \\
Ours              & 0.831 & \textbf{0.147} & \textbf{9.046} & \textbf{0.833} & \textbf{0.920} & \textbf{0.949} \\
\bottomrule
\end{tabular}
\caption{Depth estimation results on Cityscapes-style generated videos.}
\label{tab:depth_metrics_cityscapes}
\end{table}

\begin{table}[!t]
\centering
\small
\setlength{\tabcolsep}{6pt}
\renewcommand{\arraystretch}{0.85}
\begin{tabular}{lc}
\toprule
Method & mIoU \\
\midrule
Simulator only~\cite{sun2022shift} & 0.344 \\
Animatediff~\cite{guo2023animatediff} & 0.227 \\
ControlVideo~\cite{zhang2024controlvideo} & 0.258 \\
Carla2Cityscapes~\cite{pasios2025carla2real} & 0.353 \\
Wan~\cite{wan2025wan} & 0.106 \\
Ours & \textbf{0.455} \\
\bottomrule
\end{tabular}
\caption{Semantic segmentation evaluation results produced by DeepLabV3 models trained on generated data from different methods and on simulator-only data.}
\label{tab:miou_cityscapes}
\end{table}

\begin{figure}[!t]
    \centering
    \includegraphics[width=0.8\textwidth]{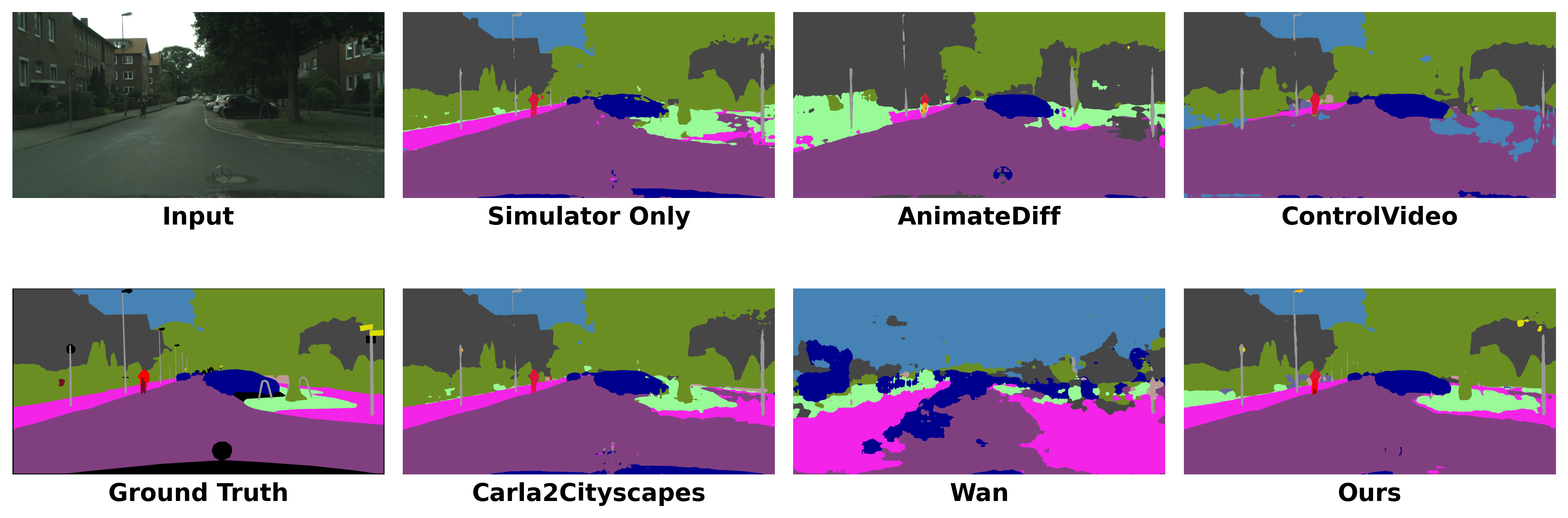}
    \caption{Visualisation of semantic segmentation results produced by DeepLabV3 models trained on generated data from different methods and on simulator-only data}
    \label{fig:seg}
\end{figure}

\subsection{Quantitative Analysis}

\noindent\textbf{DVRS evaluation results.} Table~\ref{tab:dvrs_combined} shows that our method consistently achieves the best DVRS on both the Cityscapes-style and Waymo-style settings. 
% The weighting strategy used for DVRS computation is described in Section~\ref{sec:evaluation_protocol}.
Compared with the competing methods, DriveCtrl produces generated videos that are closer to the corresponding real-world datasets not only in overall score, but also across all sub-dimensions. This trend is consistent across both target domains, indicating that the proposed method narrows the realism gap in a more comprehensive manner rather than improving only a single aspect of generation quality. Moreover, the Cityscapes and Waymo intra-domain gaps are obtained by measuring the distance between two disjoint real subsets from the corresponding real dataset. Their low values, i.e., 0.99 on Cityscapes and 3.34 on Waymo, indicate that DVRS remains relatively stable under scene-content variation within the same domain, and therefore better reflects the realism gap than generic feature-distribution metrics.

\noindent\textbf{VBench scores.} The VBench results in Table~\ref{tab:vbench_combined} provide complementary evidence from standard video-quality metrics. Our method consistently ranks best or jointly best across the evaluated dimensions. These results indicate that the proposed structure-aware adapter improves the spatial and temporal quality of generated driving videos, leading to smoother motion and fewer flickering artifacts than the compared approaches.

\noindent\textbf{Depth perception tasks.} As shown in Table~\ref{tab:depth_metrics_cityscapes}, our method achieves the best performance on most depth accuracy metrics. This suggests that the videos generated by our method better preserve the geometric information in the input depth. At the same time, our method achieves a higher SSIM than Wan, indicating that the proposed adapter alleviates the structural distortion issue in the original Wan outputs. However, it still does not achieve the highest SSIM, which suggests that some structure changes remain in the generated videos.

\noindent\textbf{Segmentation perception tasks.} As shown in Table~\ref{tab:miou_cityscapes}, when used to train DeepLabV3 for semantic segmentation, the Cityscapes-style videos generated by our method yield the best mIoU on the real Cityscapes validation set, outperforming both simulator-only training and all compared generation baselines. This result shows that the advantage of DriveCtrl is not limited to perceptual realism, but also leads to more effective synthetic training data for real-world driving perception. It further proves that the dataset constructed by our pipeline has the potential to serve as a scalable supplement to real-world datasets for improving perception algorithms in autonomous driving.

\begin{table}[!t]
\centering
\small
\setlength{\tabcolsep}{4pt}
\renewcommand{\arraystretch}{0.85}
\begin{tabular}{lccc}
\toprule
\textbf{Metric} 
& \textbf{w/o LoRA} 
& \textbf{w/o Random Reference Image} 
& \textbf{Ours} \\
\midrule
DVRS $\downarrow$          & 13.42 & 12.53  & \textbf{10.63} \\
VBench Avg. $\uparrow$  & 0.975 & 0.977  & \textbf{0.978} \\
mIoU $\uparrow$            & 0.106 & 0.203  & \textbf{0.455} \\
SSIM $\uparrow$            & 0.777 & 0.814 & \textbf{0.831} \\
\bottomrule
\end{tabular}
\caption{Ablation study of the proposed DriveCtrl framework on the Cityscape-style setting.}
\label{tab:ablation}
\end{table}

\noindent\textbf{Ablation study.} Table~\ref{tab:ablation} shows that both LoRA adaptation and random reference-image conditioning contribute to the final performance. Removing either component leads to consistent degradation, while the full model achieves the best results on all metrics.

\section{Discussion}

The results suggest that effective sim-to-real driving video generation requires balancing visual realism with structural and temporal fidelity, rather than improving appearance alone. DriveCtrl consistently achieves stronger performance across DVRS, VBench, and downstream perception tasks, indicating that the proposed structure-aware adaptation helps the model translate simulator videos toward the target real domain while better preserving scene geometry and motion patterns. The ablation results further support this interpretation, showing that both the adapter design and random reference-image conditioning are important for reducing the model's dependence on the reference image and encouraging generation to follow the condition video more faithfully. Overall, these findings highlight that realistic labelled driving video synthesis should be evaluated not only in terms of perceptual quality, but also in terms of structural faithfulness, driving-specific realism, and downstream utility.

\section{Conclusion}
In this paper, we presented \textbf{\textit{DriveCtrl}}, a depth-conditioned sim-to-real driving video generation framework that addresses the fundamental challenge of scalable realistic labelled data synthesis for autonomous driving. Built upon a pretrained video foundation model, DriveCtrl introduces a structure-aware adapter that enables controllable, depth-guided generation while faithfully preserving scene layout and motion patterns from source simulations, thereby producing temporally coherent driving videos that remain structurally aligned with the original simulated sequences. Building on this, we developed a scalable annotation-preserving sim-to-real pipeline that converts simulator video datasets into realistic driving footage matching the visual style of any target real-world dataset, with frame-wise annotations directly inherited for downstream perception tasks. To support rigorous evaluation of this task, we further proposed \textbf{\textit{DVRS}}, a driving-domain-specific world-knowledge-informed metric that holistically assesses generated video realism across plausibility, temporal coherence, structural consistency, and perceptual quality. Extensive experiments demonstrate that DriveCtrl consistently outperforms competing approaches in realism, temporal quality, and downstream perception performance, establishing a practical and scalable solution for narrowing the sim-to-real gap in driving video generation.
Nevertheless, the current study is limited to daytime driving scenarios. Future work will explore extending the framework to more challenging conditions, as well as further investigating how DriveCtrl-generated data can enhance perception-task performance in real-world deployment.

\bibliographystyle{unsrtnat}
\bibliography{reference}

@article{ye2023ip,
  title={Ip-adapter: Text compatible image prompt adapter for text-to-image diffusion models},
  author={Ye, Hu and Zhang, Jun and Liu, Sibo and Han, Xiao and Yang, Wei},
  journal={arXiv preprint arXiv:2308.06721},
  year={2023}
}

@inproceedings{mallya2020world,
  title={World-consistent video-to-video synthesis},
  author={Mallya, Arun and Wang, Ting-Chun and Sapra, Karan and Liu, Ming-Yu},
  booktitle={European Conference on Computer Vision},
  pages={359--378},
  year={2020},
  organization={Springer}
}

@inproceedings{liang2024flowvid,
  title={Flowvid: Taming imperfect optical flows for consistent video-to-video synthesis},
  author={Liang, Feng and Wu, Bichen and Wang, Jialiang and Yu, Licheng and Li, Kunpeng and Zhao, Yinan and Misra, Ishan and Huang, Jia-Bin and Zhang, Peizhao and Vajda, Peter and others},
  booktitle={Proceedings of the IEEE/CVF Conference on Computer Vision and Pattern Recognition},
  pages={8207--8216},
  year={2024}
}

@inproceedings{zhao2024exploring,
  title={Exploring generative ai for sim2real in driving data synthesis},
  author={Zhao, Haonan and Wang, Yiting and Bashford-Rogers, Thomas and Donzella, Valentina and Debattista, Kurt},
  booktitle={2024 IEEE Intelligent Vehicles Symposium (IV)},
  pages={3071--3077},
  year={2024},
  organization={IEEE}
}

@inproceedings{caesar2020nuscenes,
  title     = {nuScenes: A Multimodal Dataset for Autonomous Driving},
  author    = {Caesar, Holger and Bankiti, Varun and Lang, Alex H. and Vora, Sourabh and Liong, Venice Erin and Xu, Qiang and Krishnan, Anush and Pan, Yu and Baldan, Giancarlo and Beijbom, Oscar},
  booktitle = {Proceedings of the IEEE/CVF Conference on Computer Vision and Pattern Recognition (CVPR)},
  year      = {2020},
}

@article{chen2023controlavideo,
  title   = {Control-A-Video: Controllable Text-to-Video Diffusion Models with Motion Prior and Reward Feedback Learning},
  author  = {Chen, Weifeng and Ji, Yatai and Wu, Jie and Wu, Hefeng and Xie, Pan and Li, Jiashi and Xia, Xin and Xiao, Xuefeng and Lin, Liang},
  journal = {arXiv preprint arXiv:2305.13840},
  year    = {2023},
}

@inproceedings{cordts2016cityscapes,
  title={The cityscapes dataset for semantic urban scene understanding},
  author={Cordts, Marius and Omran, Mohamed and Ramos, Sebastian and Rehfeld, Timo and Enzweiler, Markus and Benenson, Rodrigo and Franke, Uwe and Roth, Stefan and Schiele, Bernt},
  booktitle={Proceedings of the IEEE conference on computer vision and pattern recognition},
  pages={3213--3223},
  year={2016}
}

@article{gao2024magicdrive,
  title   = {MagicDrive: Street View Generation with Diverse 3D Geometry Control},
  author  = {Gao, Ruiyuan and Chen, Kai and Xie, Enze and Hong, Lanqing and Li, Zhenguo and Yeung, Dit-Yan and Xu, Qiang},
  journal = {arXiv preprint arXiv:2310.02601},
  year    = {2024},
}

@inproceedings{gao2024vista,
  title     = {Vista: A Generalizable Driving World Model with High Fidelity and Versatile Controllability},
  author    = {Gao, Shenyuan and Yang, Jiazhi and Chen, Li and Chitta, Kashyap and Qiu, Yihang and Geiger, Andreas and Zhang, Jun and Li, Hongyang},
  booktitle = {Advances in Neural Information Processing Systems (NeurIPS)},
  year      = {2024},
}

@article{guo2023animatediff,
  title={Animatediff: Animate your personalized text-to-image diffusion models without specific tuning},
  author={Guo, Yuwei and Yang, Ceyuan and Rao, Anyi and Liang, Zhengyang and Wang, Yaohui and Qiao, Yu and Agrawala, Maneesh and Lin, Dahua and Dai, Bo},
  journal={arXiv preprint arXiv:2307.04725},
  year={2023}
}

@article{ho2022videodiffusion,
  title   = {Video Diffusion Models},
  author  = {Ho, Jonathan and Salimans, Tim and Gritsenko, Alexey and Chan, William and Norouzi, Mohammad and Fleet, David J.},
  journal = {arXiv preprint arXiv:2204.03458},
  year    = {2022},
}

@article{hu2023gaia1,
  title   = {GAIA-1: A Generative World Model for Autonomous Driving},
  author  = {Hu, Anthony and Russell, Lloyd and Yeo, Hudson and Murez, Zak and Fedoseev, George and Kendall, Alex and Shotton, Jamie and Corrado, Gianluca},
  journal = {arXiv preprint arXiv:2309.17080},
  year    = {2023},
}

@inproceedings{huang2024vbench,
  title={Vbench: Comprehensive benchmark suite for video generative models},
  author={Huang, Ziqi and He, Yinan and Yu, Jiashuo and Zhang, Fan and Si, Chenyang and Jiang, Yuming and Zhang, Yuanhan and Wu, Tianxing and Jin, Qingyang and Chanpaisit, Nattapol and others},
  booktitle={Proceedings of the IEEE/CVF Conference on Computer Vision and Pattern Recognition},
  pages={21807--21818},
  year={2024}
}

@inproceedings{kar2019metasim,
  title     = {Meta-Sim: Learning to Generate Synthetic Datasets},
  author    = {Kar, Amlan and Prakash, Aayush and Liu, Ming-Yu and Cameracci, Eric and Yuan, Justin and Rusiniak, Matt and Acuna, David and Torralba, Antonio and Fidler, Sanja},
  booktitle = {Proceedings of the IEEE/CVF International Conference on Computer Vision (ICCV)},
  year      = {2019},
}

@inproceedings{li2024drivingdiffusion,
  title     = {DrivingDiffusion: Layout-Guided Multi-View Driving Scenarios Video Generation with Latent Diffusion Model},
  author    = {Li, Xiaofan and Zhang, Yifu and Ye, Xiaoqing},
  booktitle = {European Conference on Computer Vision (ECCV)},
  year      = {2024},
}

@inproceedings{lu2024wovogen,
  title     = {WoVoGen: World Volume-aware Diffusion for Controllable Multi-camera Driving Scene Generation},
  author    = {Lu, Jiachen and Huang, Ze and Yang, Zeyu and Zhang, Jiahui and Zhang, Li},
  booktitle = {European Conference on Computer Vision (ECCV)},
  year      = {2024},
}

@article{mao2025dreamdrive,
  title   = {DreamDrive: Generative 4D Scene Modeling from Street View Images},
  author  = {Mao, Jiageng and Li, Boyi and Ivanovic, Boris and Chen, Yuxiao and Wang, Yan and You, Yurong and Xiao, Chaowei and Xu, Danfei and Pavone, Marco and Wang, Yue},
  journal = {arXiv preprint arXiv:2501.00601},
  year    = {2025},
}

@article{niu2025wise,
  title   = {WISE: A World Knowledge-Informed Semantic Evaluation for Text-to-Image Generation},
  author  = {Niu, Yuwei and Ning, Munan and Zheng, Mengren and Jin, Weiyang and Lin, Bin and Jin, Peng and Liao, Jiaqi and Feng, Chaoran and Ning, Kunpeng and Zhu, Bin and Yuan, Li},
  journal = {arXiv preprint arXiv:2503.07265},
  year    = {2025},
}

@article{pasios2025carla2real,
  title={Carla2real: A tool for reducing the sim2real appearance gap in carla simulator},
  author={Pasios, Stefanos and Nikolaidis, Nikos},
  journal={IEEE Transactions on Intelligent Transportation Systems},
  year={2025},
  publisher={IEEE}
}

@article{richter2021epe,
  title   = {Enhancing Photorealism Enhancement},
  author  = {Richter, Stephan R. and Abu Alhaija, Hassan and Koltun, Vladlen},
  journal = {arXiv preprint arXiv:2105.04619},
  year    = {2021},
}

@inproceedings{sun2020waymo,
  title     = {Scalability in Perception for Autonomous Driving: Waymo Open Dataset},
  author    = {Sun, Pei and Kretzschmar, Henrik and Dotiwalla, Xerxes and Chouard, Aur{\'e}lien and Patnaik, Vijaysai and Tsui, Paul and Guo, James and Zhou, Yin and Chai, Yuning and Caine, Benjamin and Vasudevan, Vijay and Han, Wei and Ngiam, Jiquan and Zhao, Hang and Timofeev, Aleksei and Ettinger, Scott and Krivokon, Maxim and Gao, Amy and Joshi, Aditya and Zhao, Sheng and Cheng, Shuyang and Zhang, Yu and Shlens, Jonathon and Chen, Zhifeng and Anguelov, Dragomir},
  booktitle = {Proceedings of the IEEE/CVF Conference on Computer Vision and Pattern Recognition (CVPR)},
  year      = {2020},
}

@inproceedings{sun2022shift,
  title     = {SHIFT: A Synthetic Driving Dataset for Continuous Multi-Task Domain Adaptation},
  author    = {Sun, Tao and Segu, Mattia and Postels, Janis and Wang, Yuxuan and Van Gool, Luc and Schiele, Bernt and Tombari, Federico and Yu, Fisher},
  booktitle = {Proceedings of the IEEE/CVF Conference on Computer Vision and Pattern Recognition (CVPR)},
  year      = {2022},
}

@inproceedings{wang2024drivedreamer,
  title     = {DriveDreamer: Towards Real-world-driven World Models for Autonomous Driving},
  author    = {Wang, Xiaofeng and Zhu, Zheng and Huang, Guan and Chen, Xinze and Zhu, Jiagang and Lu, Jiwen},
  booktitle = {European Conference on Computer Vision (ECCV)},
  year      = {2024},
}

@inproceedings{wang2024drivewm,
  title     = {Driving into the Future: Multiview Visual Forecasting and Planning with World Model for Autonomous Driving},
  author    = {Wang, Yuqi and He, Jiawei and Fan, Lue and Li, Hongxin and Chen, Yuntao and Zhang, Zhaoxiang},
  booktitle = {Proceedings of the IEEE/CVF Conference on Computer Vision and Pattern Recognition (CVPR)},
  year      = {2024},
}

@inproceedings{wang2024motionctrl,
  title     = {MotionCtrl: A Unified and Flexible Motion Controller for Video Generation},
  author    = {Wang, Zhouxia and Yuan, Ziyang and Wang, Xintao and Chen, Tianshui and Xia, Menghan and Luo, Ping and Shan, Ying},
  booktitle = {ACM SIGGRAPH 2024 Conference Papers},
  year      = {2024},
}

@article{wrenninge2018synscapes,
  title   = {Synscapes: A Photorealistic Synthetic Dataset for Street Scene Parsing},
  author  = {Wrenninge, Magnus and Unger, Jonas},
  journal = {arXiv preprint arXiv:1810.08705},
  year    = {2018},
}

@inproceedings{yang2020surfelgan,
  title     = {SurfelGAN: Synthesizing Realistic Sensor Data for Autonomous Driving},
  author    = {Yang, Zhenpei and Chai, Yuning and Anguelov, Dragomir and Zhou, Yin and Sun, Pei and Erhan, Dumitru and Rafferty, Sean and Kretzschmar, Henrik},
  booktitle = {Proceedings of the IEEE/CVF Conference on Computer Vision and Pattern Recognition (CVPR)},
  year      = {2020},
}

@article{yang2024cogvideox,
  title={Cogvideox: Text-to-video diffusion models with an expert transformer},
  author={Yang, Zhuoyi and Teng, Jiayan and Zheng, Wendi and Ding, Ming and Huang, Shiyu and Xu, Jiazheng and Yang, Yuanming and Hong, Wenyi and Zhang, Xiaohan and Feng, Guanyu and others},
  journal={arXiv preprint arXiv:2408.06072},
  year={2024}
}

@article{yin2023dragnuwa,
  title   = {DragNUWA: Fine-grained Control in Video Generation by Integrating Text, Image, and Trajectory},
  author  = {Yin, Shengming and Wu, Chenfei and Liang, Jian and Shi, Jie and Li, Houqiang and Gong, Ming and Duan, Nan},
  journal = {arXiv preprint arXiv:2308.08089},
  year    = {2023},
}

@inproceedings{yu2020bdd100k,
  title     = {BDD100K: A Diverse Driving Dataset for Heterogeneous Multitask Learning},
  author    = {Yu, Fisher and Chen, Haofeng and Wang, Xin and Xian, Wenqi and Chen, Yingying and Liu, Fangchen and Madhavan, Vashisht and Darrell, Trevor},
  booktitle = {Proceedings of the IEEE/CVF Conference on Computer Vision and Pattern Recognition (CVPR)},
  year      = {2020},
}

@inproceedings{zhang2023adding,
  title={Adding conditional control to text-to-image diffusion models},
  author={Zhang, Lvmin and Rao, Anyi and Agrawala, Maneesh},
  booktitle={Proceedings of the IEEE/CVF international conference on computer vision},
  pages={3836--3847},
  year={2023}
}

@inproceedings{zhang2024controlvideo,
  title     = {ControlVideo: Training-Free Controllable Text-to-Video Generation},
  author    = {Zhang, Yabo and Wei, Yuxiang and Jiang, Dongsheng and Zhang, Xiaopeng and Zuo, Wangmeng and Tian, Qi},
  booktitle = {International Conference on Learning Representations (ICLR)},
  year      = {2024},
}

@article{zhang2025worldgenbench,
  title   = {WorldGenBench: A World-Knowledge-Integrated Benchmark for Reasoning-Driven Text-to-Image Generation},
  author  = {Zhang, Daoan and Jiang, Che and Xu, Ruoshi and Chen, Biaoxiang and Jin, Zijian and Lu, Yutian and Zhang, Jianguo and Yong, Liang and Luo, Jiebo and Luo, Shengda},
  journal = {arXiv preprint arXiv:2505.01490},
  year    = {2025},
}

@inproceedings{zhao2025drivedreamer2,
  title     = {DriveDreamer-2: LLM-Enhanced World Models for Diverse Driving Video Generation},
  author    = {Zhao, Guosheng and Wang, Xiaofeng and Zhu, Zheng and Chen, Xinze and Huang, Guan and Bao, Xiaoyi and Wang, Xingang},
  booktitle = {Proceedings of the AAAI Conference on Artificial Intelligence},
  year      = {2025},
}

@article{zheng2025vbench2,
  title   = {VBench-2.0: Advancing Video Generation Benchmark Suite for Intrinsic Faithfulness},
  author  = {Zheng, Dian and Huang, Ziqi and Liu, Hongbo and Zou, Kai and He, Yinan and Zhang, Fan and Gu, Lulu and Zhang, Yuanhan and He, Jingwen and Zheng, Wei-Shi and Qiao, Yu and Liu, Ziwei},
  journal = {arXiv preprint arXiv:2503.21755},
  year    = {2025},
}

@inproceedings{heusel2017gans,
  title={GANs Trained by a Two Time-Scale Update Rule Converge to a Local Nash Equilibrium},
  author={Heusel, Martin and Ramsauer, Hubert and Unterthiner, Thomas and Nessler, Bernhard and Hochreiter, Sepp},
  booktitle={Advances in Neural Information Processing Systems},
  volume={30},
  pages={6626--6637},
  year={2017}
}

@article{unterthiner2018fvd,
  title={Towards Accurate Generative Models of Video: A New Metric \& Challenges},
  author={Unterthiner, Thomas and van Steenkiste, Sjoerd and Kurach, Karol and Marinier, Raphael and Michalski, Marcin and Gelly, Sylvain},
  journal={arXiv preprint arXiv:1812.01717},
  year={2018}
}

@article{nvidia2025cosmostransfer,
  title={World Simulation with Video Foundation Models for Physical AI},
  author={{NVIDIA} and Ali, Arslan and others},
  journal={arXiv preprint arXiv:2511.00062},
  year={2025}
}

@article{xi2025omnivdiff,
  title={OmniVDiff: Omni Controllable Video Diffusion for Generation and Understanding},
  author={Xi, Dianbing and Wang, Jiepeng and Liang, Yuanzhi and Qi, Xi and Huo, Yuchi and Wang, Rui and Zhang, Chi and Li, Xuelong},
  journal={arXiv preprint arXiv:2504.10825},
  year={2025}
}

@article{wan2025wan,
  title={Wan: Open and advanced large-scale video generative models},
  author={Wan, Team and Wang, Ang and Ai, Baole and Wen, Bin and Mao, Chaojie and Xie, Chen-Wei and Chen, Di and Yu, Feiwu and Zhao, Haiming and Yang, Jianxiao and others},
  journal={arXiv preprint arXiv:2503.20314},
  year={2025}
}

@article{chen2017deeplabv3,
  title={Rethinking Atrous Convolution for Semantic Image Segmentation},
  author={Chen, Liang-Chieh and Papandreou, George and Schroff, Florian and Adam, Hartwig},
  journal={arXiv preprint arXiv:1706.05587},
  year={2017}
}

@article{yang2024depth,
  title={Depth anything v2},
  author={Yang, Lihe and Kang, Bingyi and Huang, Zilong and Zhao, Zhen and Xu, Xiaogang and Feng, Jiashi and Zhao, Hengshuang},
  journal={Advances in Neural Information Processing Systems},
  volume={37},
  pages={21875--21911},
  year={2024}
}

@inproceedings{yang2023unisim,
  title={Unisim: A neural closed-loop sensor simulator},
  author={Yang, Ze and Chen, Yun and Wang, Jingkang and Manivasagam, Sivabalan and Ma, Wei-Chiu and Yang, Anqi Joyce and Urtasun, Raquel},
  booktitle={Proceedings of the IEEE/CVF Conference on Computer Vision and Pattern Recognition},
  pages={1389--1399},
  year={2023}
}

@inproceedings{chen2024geodiffusion,
  title={{GeoDiffusion}: Text-Prompted Geometric Control for Object Detection Data Generation},
  author={Kai Chen and Enze Xie and Zhe Chen and Yibo Wang and Lanqing Hong and Zhenguo Li and Dit-Yan Yeung},
  booktitle={International Conference on Learning Representations (ICLR)},
  year={2024}
}

@inproceedings{wen2024panacea,
  title={Panacea: Panoramic and Controllable Video Generation for Autonomous Driving},
  author={Yuqing Wen and Yucheng Zhao and Yingfei Liu and Fan Jia and Yanhui Wang and Chong Luo and Chi Zhang and Tiancai Wang and Xiaoyan Sun and Xiangyu Zhang},
  booktitle={Proceedings of the IEEE/CVF Conference on Computer Vision and Pattern Recognition (CVPR)},
  pages={6902--6912},
  year={2024}
}

@article{wen2024panaceaplus,
  title={Panacea+: Panoramic and Controllable Video Generation for Autonomous Driving},
  author={Yuqing Wen and Yucheng Zhao and Yingfei Liu and Binyuan Huang and Fan Jia and Yanhui Wang and Chi Zhang and Tiancai Wang and Xiaoyan Sun and Xiangyu Zhang},
  journal={arXiv preprint arXiv:2408.07605},
  year={2024}
}

@article{gao2024magicdrive3d,
  title={{MagicDrive3D}: Controllable 3D Generation for Any-View Rendering in Street Scenes},
  author={Ruiyuan Gao and Kai Chen and Zhihao Li and Lanqing Hong and Zhenguo Li and Qiang Xu},
  journal={arXiv preprint arXiv:2405.14475},
  year={2024}
}

@article{bansal2024videophy,
  title={Videophy: Evaluating physical commonsense for video generation},
  author={Bansal, Hritik and Lin, Zongyu and Xie, Tianyi and Zong, Zeshun and Yarom, Michal and Bitton, Yonatan and Jiang, Chenfanfu and Sun, Yizhou and Chang, Kai-Wei and Grover, Aditya},
  journal={arXiv preprint arXiv:2406.03520},
  year={2024}
}

@article{bansal2025videophy,
  title={Videophy-2: A challenging action-centric physical commonsense evaluation in video generation},
  author={Bansal, Hritik and Peng, Clark and Bitton, Yonatan and Goldenberg, Roman and Grover, Aditya and Chang, Kai-Wei},
  journal={arXiv preprint arXiv:2503.06800},
  year={2025}
}

@article{meng2024towards,
  title={Towards world simulator: Crafting physical commonsense-based benchmark for video generation},
  author={Meng, Fanqing and Liao, Jiaqi and Tan, Xinyu and Shao, Wenqi and Lu, Quanfeng and Zhang, Kaipeng and Cheng, Yu and Li, Dianqi and Qiao, Yu and Luo, Ping},
  journal={arXiv preprint arXiv:2410.05363},
  year={2024}
}

@inproceedings{chen2026t2vworldbench,
  title={T2vworldbench: A benchmark for evaluating world knowledge in text-to-video generation},
  author={Chen, Yubin and Guo, Xuyang and Shi, Zhenmei and Song, Zhao and Zhang, Jiahao},
  booktitle={Proceedings of the IEEE/CVF Winter Conference on Applications of Computer Vision},
  pages={6474--6485},
  year={2026}
}

@article{hu2022lora,
  title={Lora: Low-rank adaptation of large language models.},
  author={Hu, Edward J and Shen, Yelong and Wallis, Phillip and Allen-Zhu, Zeyuan and Li, Yuanzhi and Wang, Shean and Wang, Liang and Chen, Weizhu and others},
  journal={Iclr},
  volume={1},
  number={2},
  pages={3},
  year={2022}
}

\end{document}